\documentclass[journal]{IEEEtran}

\usepackage{times}
\usepackage{mathrsfs}
\usepackage{epsfig}
\usepackage{graphicx}
\usepackage{amsmath}
\usepackage{amssymb}
\usepackage{graphicx}
\usepackage{cite}
\usepackage{enumerate}
\usepackage{cases}
\usepackage{multirow}
\usepackage{verbatim}
\usepackage{amssymb}
\usepackage{CJK}
\usepackage{algorithm}
\usepackage{algorithmicx}
\usepackage{algpseudocode}
\usepackage{color}
\usepackage{bm}
\usepackage{booktabs}
\usepackage[colorlinks,linkcolor=red]{hyperref}
\usepackage{amssymb}
\usepackage[frenchb]{babel}
\usepackage[utf8]{inputenc}
\usepackage[T1]{fontenc}

\newcommand{\etal}{\textit{et al}.}
\newcommand{\ie}{\textit{i}.\textit{e}.}
\newcommand{\eg}{\textit{e}.\textit{g}.}
\newcommand{\etc}{\textit{etc}.}

\begin{document}

\title{RRNet: Relational Reasoning Network with Parallel Multi-scale Attention for Salient Object Detection in Optical Remote Sensing Images}

\author
{
Runmin Cong,~\IEEEmembership{Member,~IEEE,} Yumo Zhang, Leyuan Fang,~\IEEEmembership{Senior Member,~IEEE,} Jun Li,~\IEEEmembership{Fellow,~IEEE,}\\ Yao Zhao,~\IEEEmembership{Senior Member,~IEEE,} and Sam Kwong,~\IEEEmembership{Fellow,~IEEE}

\thanks{R. Cong is with the Institute of Information Science, Beijing Jiaotong University, Beijing 100044, China, also with the Beijing Key Laboratory of Advanced Information Science and Network Technology, Beijing 100044, China, and also with the Department of Computer Science, City University of Hong Kong, Hong Kong SAR, China (e-mail: rmcong@bjtu.edu.cn).}
\thanks{Y. Zhang and Y. Zhao are with the Institute of Information Science, Beijing Jiaotong University, Beijing 100044, China, and also with the Beijing Key Laboratory of Advanced Information Science and Network Technology, Beijing 100044, China (e-mail: yumozhang@bjtu.edu.cn; yzhao@bjtu.edu.cn).}
\thanks{L. Fang is with the College of Electrical and Information Engineering, Hunan University, Changsha 410082, China, and also with the Peng Cheng Laboratory, Shenzhen 518000, China (e-mail: fangleyuan@gmail.com).}
\thanks{J. Li is with the Guangdong Provincial Key Laboratory of Urbanization and Geosimulation, School of Geography and Planning, Sun Yat-sen University, Guangzhou 510275, China (e-mail: lijun48@mail.sysu.edu.cn).}
\thanks{S. Kwong is with the Department of Computer Science, City University of Hong Kong, Hong Kong SAR, China, and also with the City University of Hong Kong Shenzhen Research Institute, Shenzhen 51800, China (e-mail: cssamk@cityu.edu.hk).}
}

\markboth{IEEE TRANSACTIONS ON GEOSCIENCE AND REMOTE SENSING}
{Shell \MakeLowercase{\textit{et al.}}: Bare Demo of IEEEtran.cls for IEEE Journals}
\maketitle

\begin{abstract}
Salient object detection (SOD) for optical remote sensing images (RSIs) aims at locating and extracting visually distinctive objects/regions from the optical RSIs. Despite some saliency models were proposed to solve the intrinsic problem of optical RSIs (such as complex background and scale-variant objects), the accuracy and completeness are still unsatisfactory. To this end, we propose a relational reasoning network with parallel multi-scale attention for SOD in optical RSIs in this paper. The relational reasoning module that integrates the spatial and the channel dimensions is designed to infer the semantic relationship by utilizing high-level encoder features, thereby promoting the generation of more complete detection results. The parallel multi-scale attention module is proposed to effectively restore the detail information and address the scale variation of salient objects by using the low-level features refined by multi-scale attention. Extensive experiments on two datasets demonstrate that our proposed RRNet outperforms the existing state-of-the-art SOD competitors both qualitatively and quantitatively. \url{https://rmcong.github.io/proj\_RRNet.html}.
\end{abstract}


\IEEEpeerreviewmaketitle

\section{Introduction} \label{sec1}
\IEEEPARstart{H}{uman} beings' attention can be easily grabbed by the distinctive objects/regions in an image \cite{crm_review}. Imitating this visual attention system, salient object detection (SOD) aims at accurately locating the most attractive objects/regions, which has been widely applied in many tasks, due to its efficiency and convenience, such as segmentation \cite{intro-3}, retargeting \cite{intro-1}, thumbnail creation \cite{intro-4}, visual quality assessment \cite{intro-6}, traffic sign detection\cite{yuan2019vssa, wang2020multitask}, and group detection\cite{li2017multiview}, \etc.  In addition to the common SOD methods for natural scene images (NSIs) \cite{R3Net, RADF, PFAN, MINet,GateNet,F3Net,EGNet,PoolNet,GCPANet}, the SOD family includes many members for different data requirements, such as RGB-D SOD \cite{RGBD-1,RGBD-3,crmspl,crmtc20,eccv20}, co-saliency detection \cite{nips20,crmtmm19,crmtip18,crmtc19}, video SOD \cite{Video-1}, and SOD in optical RSIs \cite{DAFNet,LVNet,lcync20}. In this paper, we focus on the SOD task in optical RSIs, which can be beneficial to many downstream remote sensing tasks, such as ROI extraction \cite{SMFF} and object detection \cite{wu2019orsim}.
\begin{figure}[!t]
\centering
\centerline{\includegraphics[width=1\linewidth]{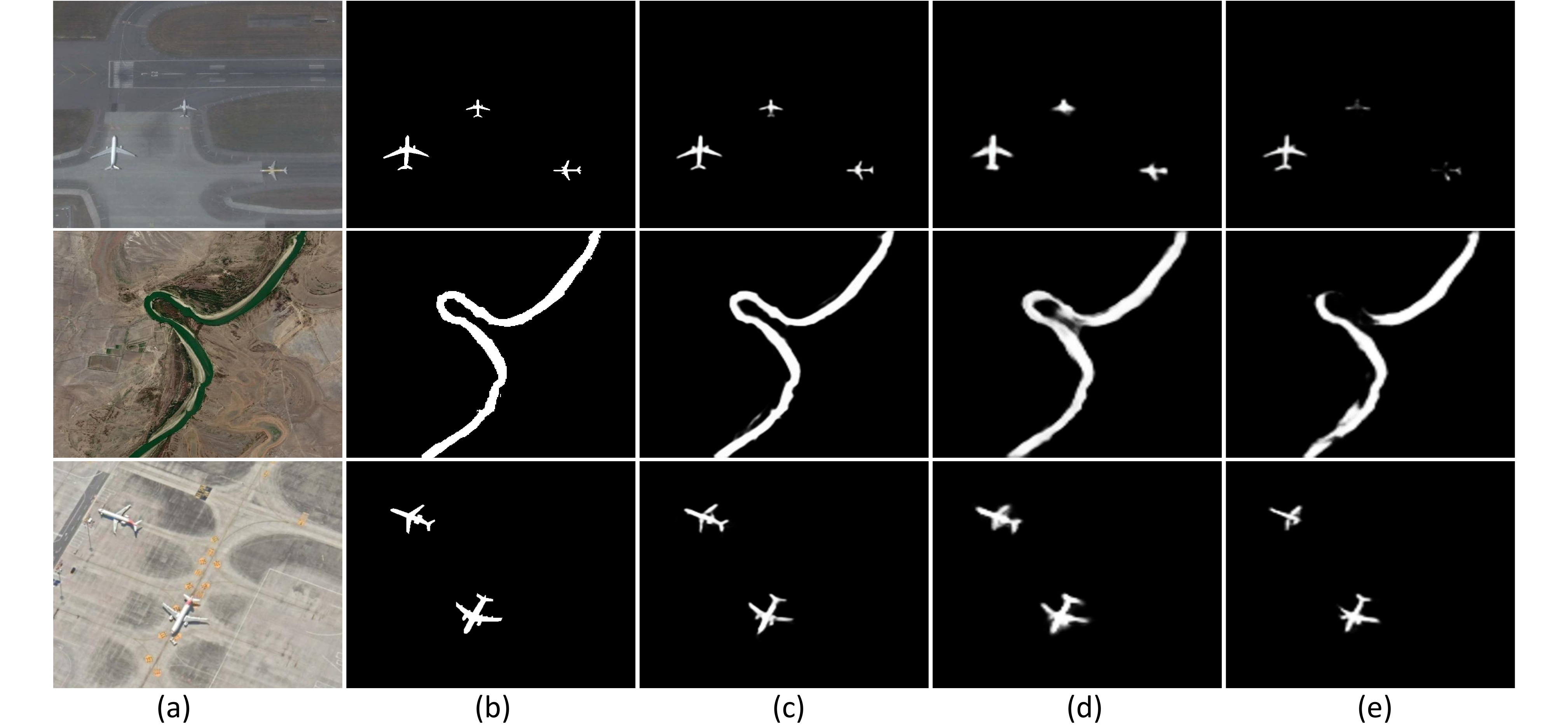}}
\caption{Visual examples of complicated scenarios in RSIs. (a) optical RSIs. (b) ground truth. (c) proposed RRNet. (d) DAFNet. (e) LVNet.}
\label{fig1}
\end{figure}
Optical RSIs in this paper refer specifically to the color images collected from airborne or satellite sources in the range of $400 \sim 760$ nm. Our naming strategy is consistent with several previous works, such as object detection \cite{li2020object, wu2019orsim}, ship detection \cite{rsi6}, region search \cite{yin2019region} and salient object detection \cite{LVNet, lcync20, DAFNet}. In these works, the airborne color images with high spatial resolution are named as the optical remote sensing images. Optical RSIs have been widely used in land planning, disaster monitoring, military reconnaissance and other fields in light of its characteristics of large range, high resolution, content richness and intuitiveness. Although NSIs and optical RSIs have similar color representation, when we directly use the state-of-the-art SOD method in NSIs to process optical RSI data, it is still difficult to obtain satisfactory results due to the different imaging environments and conditions. To be specific, unlike the NSIs captured from a handheld camera perspective, optical RSIs are obtained from a bird's-eye view of the ground on a satellite or airplane, and this process is easily affected by environment factors such as shooting altitude, time, and illumination. Therefore, many challenging problems will arise in the optical RSI SOD task.
First, there are challenges in terms of object number and scale. Affected by high-altitude shooting and large area coverage, multiple salient objects are prone to appear in the optical RSIs, which greatly increases the difficulty of detection. Moreover, the bird's-eye view also raises another problem, that is, the objects in optical RSIs show large scale variations, from the automobile occupying several pixels to a river throughout the image. Even objects belonging to the same category may have different scales due to the different imaging collection heights. In addition, different from the NSIs captured by handheld cameras, the optical RSIs acquisition process adopts a wide-range scanning-like manner, which results in the position of the salient objects being almost random, thus invalidating the conventional center prior. Second, there are challenges in complex background and imaging interference.  The outdoor overhead imaging mode and the wide imaging range make the optical RSIs often contain more complicated backgrounds. Moreover, other environment factors, such as weather condition, illumination intensity, and shooting time, may cause diverse unpredictable disturbances which manifest as shadows, over-exposure, appearance changes, and shape distortions.

In response to above issues, specially designed SOD models for optical RSIs have been introduced \cite{LVNet, DAFNet}, but there are still some issues that need to be better addressed. On one hand, missing and incomplete detection of salient objects are common problems in the existing methods. In a multi-objects scenario, salient objects may not appear concentrated, but scattered in multiple locations of the image, such as the three airplanes in the first row of Fig. \ref{fig1}. In fact, these objects are still correlated in semantic attributes. Even for a single-object scenario, similar problem may be encountered. For example, the salient object spans a larger range in the image (\eg, the river in the second row of Fig. \ref{fig1}), which makes it difficult to detect it completely. Regarding this issue, context mechanism has been implemented to address it by modeling the global relation \cite{DAFNet}. This solution can alleviate the missing and incomplete of objects by extending receptive field to a certain extent, but the semantic information is not fully utilized. Therefore, in this paper, we try to embed a relational reasoning module in the high-level encoder stage to model the semantic relation of different objects or different parts of objects on the graph. Benefiting from the representation ability of graph structure, our designed relational reasoning mechanism can not only construct the constraint relationship between each spatial region, but also reason about the semantic relationship of the channel dimension\footnote{The channel dimension refers to the channel dimension of the learned high-level features (tensor), rather than the concept of spectral channels.}, thereby obtaining a more comprehensive and deeper internal modeling relationship.

On the other hand, high-level features containing abstract semantic information are suitable for relational reasoning to achieve complete detection, while low-level features have higher spatial resolution and more detailed information (such as boundary) which can supplement the decoding process to achieve detail restoration. However, directly employing the coarse low-level features may cause additional redundancy interference, with considering that objects of different sizes at high spatial resolution are easy to detect, we propose a parallel multi-scale attention mechanism that can suppress unimportant features while restoring low-level details. In the decoder stage of the existing encoder-decoder networks, the lost details are not only difficult to restore through a sequence of deconvolution layers, but also be coarsen by the up-sampling operation, such as planes in the third row of Fig. \ref{fig1}. Therefore, we deal with low-level features by multiple receptive fields in two parallel ways. Furthermore, low-level features are refined into 2D attention maps which can restore the detail information effectively.

In summary, an end-to-end SOD model for optical RSIs is proposed which focuses on reasoning semantic relations and restoring details. The major contributions of the proposed method are summarized as follows.

\begin{itemize}
	\item[1)]
	We propose a novel end-to-end relational reasoning network with parallel multi-scale attention (RRNet) for SOD in optical RSIs, which consists of a relational reasoning encoder and a multi-scale attention decoder.
\end{itemize}
\begin{itemize}
	\item[2)]
	We design a relational reasoning module in the high-level layers of the encoder stage to model the sematic relations and force the generation of complete salient objects. This is the first attempt to introduce relational reasoning in the SOD framework for optical RSIs. Moreover, we innovatively employ relational reasoning along the spatial and channel dimensions jointly to obtain more comprehensive semantic relations.
\end{itemize}
\begin{itemize}
	\item[3)]
	We propose a parallel multi-scale attention scheme in the low-level layers of the decoder stage to recover the detail information in a multi-scale and attention manner. This mechanism can deal with the object scale variation issue through the multi-scale design, while effectively recovering the detail information with the help of shallower features selected by the parallel attention.
\end{itemize}
\begin{itemize}
	\item[4)]
	We compare the proposed methods with thirteen state-of-the-art approaches on two challenging optical RSI datasets. Without bells and whistles, our method achieves the best performance under three evaluation metrics. Besides, the model has a real-time inference speed of 109 FPS.
\end{itemize}


\section{Related Work} \label{sec2}

%

\begin{figure*}[!t]
	\centering
	\centerline{\includegraphics[width=1\linewidth]{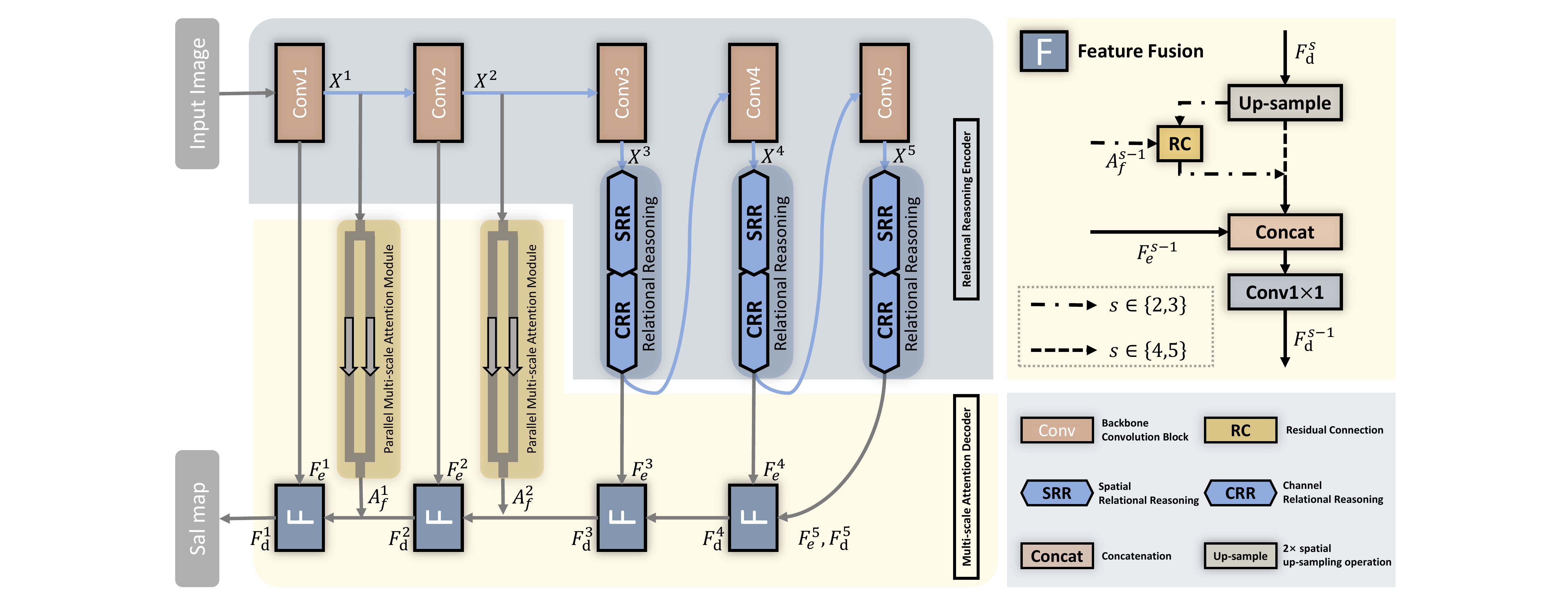}}
	\caption{The overall framework of the proposed RRNet, consisting of a relational reasoning encoder and a multi-scale attention decoder. The encoder generates hierarchical features, \ie, low-level features from the first two stages and high-level features from the last three stages. Relational reasoning in two dimensions following each high-level stage are successively employed to refine features by reasoning semantic relationship. Low-level features obtained by encoder are fed into parallel multi-scale attention module, generating attention maps with valuable information to restore lost details. The top right portion is the computation procedures of feature fusion between passed-up deep features and shallow features.}
	\label{fig2}
\end{figure*}
\subsection{Traditional SOD Models in optical RSIs}
The unique imaging conditions and environment make the salient object detection in optical RSIs more challenging, and thus some works have begun to study specialized algorithms instead of directly transplanting SOD models for NSIs into optical RSIs community, which usually fails to achieve satisfactory performance. Similar to the bottom-up SOD models in NSIs, some traditional methods have been proposed. Zhao \etal \cite{SGS} used the global and background cues to define the corresponding sparse dictionary, and calculated the saliency of the optical RSI based on sparse representation. Zhang \etal \cite{SMFF} extracted the color, intensity, texture, and global contrast features and designed a low-rank matrix recovery based feature fusion method to calculate the saliency detection in RSIs. Ma \etal \cite{ROI1} calculated the superpixel-level saliency based on the structure tensor and background contrast, and then obtained the final pixel-level saliency map by superpixel-to-pixel mapping. Zhang \etal \cite{ROI2} used the quaternion Fourier transform in the frequency domain to generate the saliency map. Zhang \etal \cite{ROI3} computed the saliency of the remote sensing images based on the multiscale visual saliency analysis, which integrates the intensity, orientation, and color features.


\subsection{Deep Learning based SOD Models in optical RSIs}
Thanks to the powerful feature representation capabilities of deep learning, related models have gradually emerged and developed.
As a pioneering work, Li \etal ~made a positive attempt in \cite{LVNet}. First, in order to make up for the lack of public datasets, the first public available dataset named ORSSD containing $600$ training and $200$ testing was constructed, which bridged the gap between the data and algorithm verification. In addition, an end-to-end SOD network for optical RSIs named LVNet is proposed, where a two-stream pyramid module is designed to address the scale variation of salient objects, and a encoder-decoder architecture with nested connections aims to learn more discriminative saliency features.
Zhang \etal \cite{DAFNet} extended the ORSSD dataset as a larger one named EORSSD dataset, which includes $1,400$ training images and $600$ testing images. Moreover, an end-to-end Dense Attention Fluid Network (DAFNet) for SOD in optical RSIs is proposed, in which a global context-aware attention module is embedded in a dense attention fluid structure to learn more discriminative saliency features with long-range context constraints.
Zhang \etal \cite{zhang2020rssod} proposed a progressively supervised learning method for SOD in RSIs, including the pseudo-label generation with auxiliary image and weakly supervised to fully supervised learning. The former module is designed to generate the pseudo SOD label by using a classification network with Grad-CAM, and the latter process designs a feedback saliency analysis network equipped with curriculum learning strategy under the supervision of the generated pseudo-label to achieve saliency detection.


\begin{figure*}[!t]
	\centering
	\includegraphics[width=1\linewidth]{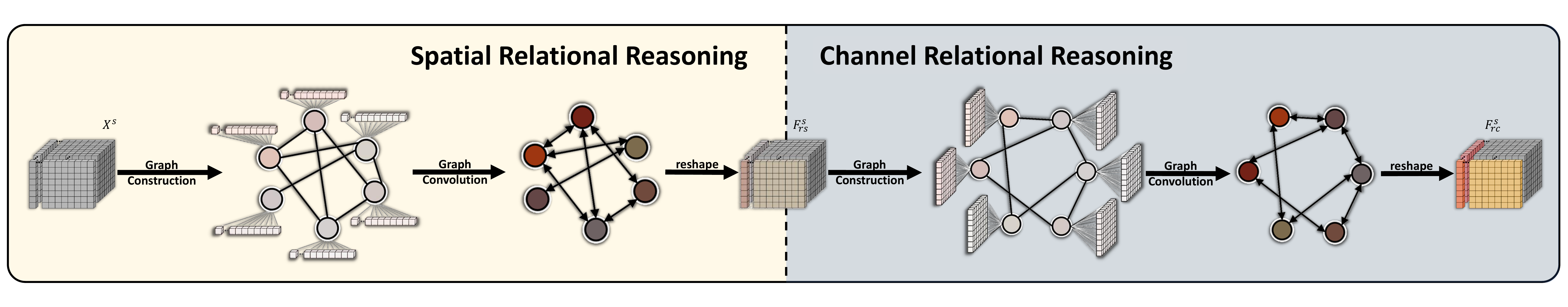}
	\caption{Illustration of the relational reasoning module, including spatial relational reasoning (the left part) and channel relational reasoning (the right part). }
	\label{fig3}
\end{figure*}
\section{Proposed Method} \label{sec3}

\subsection{Overview}
As illustrated in Fig. \ref{fig2}, we propose an encoder-decoder network, named RRNet, for SOD in optical RSIs, which models the semantic relation and restores the multi-scale details. The overall framework can be roughly divided into relation reasoning encoder stage and multi-scale attention decoder stage.

The relational reasoning encoder encodes semantic relations generated from high-level features to learn effective feature descriptions. Our backbone extractor consists of five sequentially-stacked convolutional blocks that obtain the corresponding low-level features from the first two convolutional blocks and high-level features from the last three convolutional blocks. These five convolutional block features are denoted as $\{X^1,X^2,X^3,X^4,X^5\}$. Then, each high-level convolutional features are followed by a relational reasoning module to reason and model sematic relation of objects or regions. The reasoning process is carried out on the graph model in the spatial dimension and channel dimension successively, thereby generating the enhanced features as the input tensor of the next convolutional block.
The multi-scale attention decoder progressively generates the saliency-related referring features, and restores detail information by being equipped with a parallel multi-scale attention mechanism at the low-level layers. During feature decoding, we progressively fuse the feature maps of different levels, and embed a parallel multi-scale attention module in the low-level layers to extract multi-scale attention map from two multi-scale calculation angles. Then, the obtained attention map is used to refine the passed-up fused features via residual connection. Finally, we utilize the last decoding features to predict the final saliency map through additional convolution layers.

In the following subsections, we will introduce the \emph{Relational reasoning Encoder}, \emph{Multi-scale Attention Decoder}, and \emph{Loss Function}, respectively.

\subsection{Relational Reasoning Encoder}
Because of the diverse object numbers, varying object scales, and overwhelming background context redundancies in the optical RSIs, it is difficult to detect the salient objects with accurate location and complete structure. For example, in the first image of Fig. \ref{fig1}, the multiple objects share the similar appearance and semantic attribute, once one object is determined, the relational information between the objects can be introduced to assist the detection of other objects. Analogously, the relational dependency is also existed in the interior of the object, as shown in the second image of Fig. \ref{fig1}. Different parts of the object should be highlighted uniformly, and constructing relationships between different spatial regions within the object is conducive to achieving this goal. The graph-based models have been proven to be effective for context reasoning in many tasks, such as visual recognition \cite{GR2}, human-object interaction detection \cite{GR7}. Through the graph, capturing relations between arbitrary regions in the input can be simplified to explore interactions between the features of the corresponding nodes. Thus, we introduce the relational reasoning mechanism into the encoder stage, which establishes the semantic relation between each pixel and each channel on the graph.

As the number of convolutional layers increases, the learned feature representations gradually shift from low-level information (such as texture and edge details) to high-level semantic information (such as category attributes). It is happened to be the relational reasoning process exactly needs these high-level semantic information.
Chen \etal \cite{GR6} proposed a global reasoning module which projects features from the coordinate space into an interactive space and conducts graph convolution for global reasoning. However, it only makes reasoning along the spatial space, but ignores the interaction of the features along the channel dimension. Thus, to obtain a more comprehensive and deeper internal modeling relationship, we perform graph reasoning modeling in both spatial and channel dimensions. Specifically, our relational reasoning module is conducted on the graph model, and decoupled into two different dimensions, \ie, spatial relational reasoning and channel relational reasoning. The high-level convolutional features $X^s~(s\in{\{3,4,5\}})$ are first fed to spatial relational reasoning (SRR) module, thereby generating spatial reasoning features $F_{rs}^s$. Then, the channel relational reasoning (CRR) module is applied on the features $F_{rs}^s$ to generate the channel reasoning features $F_{rc}^s$. The detailed illustration is provided in Fig. \ref{fig3}.

\subsubsection{\textbf{Graph Construction}}

For graph reasoning in the existing methods, the features in coordinate space need to be first projected into graph space to apply a general graph convolution.
Then, a reverse projection is conducted to transform the resulting features back to the original coordinate space.
Instead, we skip the projecting and re-projecting processes to spare the computation by deeming the feature space as a special case of data defined on a low-dimensional graph.
We will provide the details of spatial/channel graphs and adjacency matrix calculation.

\emph{Spatial and Channel Graphs.} How to completely detect the salient objects covering long-distance ranges is a difficult issue in SOD for optical RSIs, because the two ends of these objects often have a large span, and even pass through the entire image (such as the river in Fig. \ref{fig1}). Spatial relational reasoning can establish the semantic relationship between any two spatial locations, thereby constraining that no matter how long the range of the object is, their semantic relations are associated. For the input feature $X^s\in{ \mathbb{R} ^ { H \times W\times C }}$, we first reshape it into $G_s^s\in{\mathbb{R}^{ HW\times C}}$. Therefore, we can build the spatial graph with $HW$ vertexes, and each vertex is represented by the corresponding channel feature of $\mathbb{R}^{C \times 1}$.

In terms of the high-level features, they often contain more compact semantic information, but their spatial resolution tends to be relatively small, while channel information is relatively large and rich. If only implementing relational reasoning in the spatial space, many useful channel features will not fully utilized. Thus, we innovatively build a channel graph and apply relational reasoning on it. For the graph construction, each channel of the high-level features is regarded as a vertex, and all the spatial information of the corresponding channel is used to describe the vertex. Assuming the output features of SRR module are denoted as $F_{rs}^s\in{\mathbb{R}^{ H \times W \times C }},$ we reshape it into a 2D tensor $G_c^s\in{\mathbb{R}^{C\times HW }}$, and $C$ vertexes are further obtained, which is represented by the corresponding spatial feature of $\mathbb{R}^{HW \times 1}$. In this way, the graph convolution on the channel graph can model the semantic relationship between different channels.

\emph{Adjacency Matrix.} After determining the vertex information on the corresponding graph model $G^s\in{\mathbb{R}^{a_1 \times a_2}}$, we define the adjacency matrix $\tilde{A}$ to indicate the similarity of pair-wise vertices, where $a_1$ denotes the number of vertexes, and $a_2$ represents the feature dimensionality of each vertex. Following \cite{GR5}, we use the Euclidean distance to measure the similarity between vertex $i$ and vertex $j$ in adjacency matrix $\tilde{A}$. In implementation, dot-product distance is employed to calculate $\tilde{A}=[\tilde{A}_{ij}]\in{\mathbb{R}^{a_1 \times a_1}}$ as follows:
\begin{equation}
	\tilde{A}_{ij}= (conv_{1 \times 1}(G^s))_i \cdot \tilde{\Lambda}(G^s)\cdot(conv_{1 \times 1}(G^s))_j^T,
\end{equation}
where $conv_{1 \times 1}$ is a customized $1 \times 1$ convolutional layer followed by ReLU non-linearity activation. $\tilde{\Lambda}(G^s)$ is a diagonal matrix which has attention on the inner product to learn a better distance metric, which is defined as:
\begin{equation}
	\tilde{\Lambda}(G^s)=diag(conv_{1 \times 1}(avepool(G^s))),
\end{equation}
where $avepool(\cdot)$ is the average-pooling, and $diag(\cdot)$ reshapes a vector into a diagonal matrix.
\begin{figure}[t]
	\centering
	\includegraphics[width=1\columnwidth]{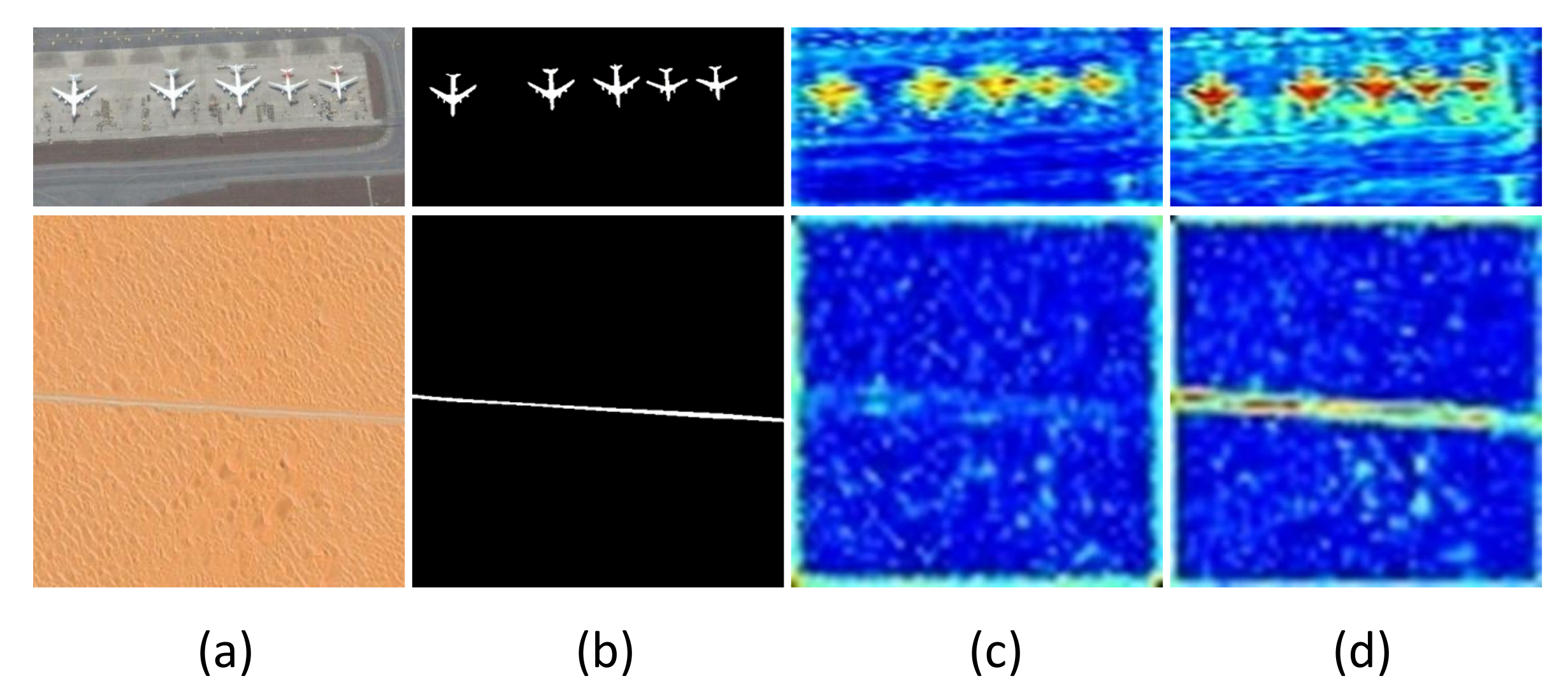}
	\caption {Visualization of features in relational reasoning module. (a) optical RSIs. (b) ground truth. (c) features fed into relational reasoning module. (d) features generated by relational reasoning module.
	}
	\label{fig4}
\end{figure}
\subsubsection{\textbf{Graph Reasoning}}

For graph information $G^s\in{\mathbb{R}^{a_1 \times a_2}}$, which is reshaped from original side feature map of last three convolution stages, $s\in\{3,4,5\}$ indexes the convolution stage. Our goal is to generate output features $F_r^s\in{\mathbb{R}^{H \times W \times C }}$ encoded by graph reasoning, which is defined as:
\begin{equation}
	F_r^s=\sigma(\tilde{L}G^s\Theta),
\end{equation}
where $\Theta$ is a trainable weight matrix, $\tilde{L}$ is the graph Laplacian matrix, and $\sigma$ is the ReLU activation function. The Laplacian matrix we used here is data-dependent parameters in order to capture the sematic relation between vertex features dynamically. Thus, $\tilde{L}$ is not restricted as a specific one, changing with the different features. It is formulated with the symmetric normalized form:
\begin{equation}
	\tilde{L}=I-\tilde{D}^{-\frac{1}{2}}\tilde{A}\tilde{D}^{-\frac{1}{2}},
\end{equation}
where $\tilde{D}=diag(d_1,d_2,...,d_n)$, $d_i=\sum_j\tilde{A}_{ij}$, $\tilde{A}\in{\mathbb{R}^{a_2 \times a_2}}$ is the data-dependent adjacency matrix, and $I$ is the unit matrix. As mentioned earlier, if we make reasoning on the spatial graph, the obtained feature is denoted as $F_{rs}^s$. If the reasoning is made on the channel graph, it is represented as $F_{rc}^s$.
The feature visualizations before and after relational reasoning are shown in Fig. \ref{fig4}. From it, we can see that after the graph reasoning module, the integrity between multiple salient objects and within the same salient target has been significantly improved.

\subsection{Multi-scale Attention Decoder}
After learning the convolutional features in the encoder stage, the decoder stage aims to gradually fuse the feature maps of different levels and generate the saliency-related features. Considering the complementary role of encoder features, the up-sampled decoder features and the corresponding encoder features are integrated by a fusion unit (as shown in the upper right corner of Fig. \ref{fig2}). As we all know, low-level features have a larger spatial resolution, which include more detail information, and is conducive to recovering the salient objects of different scale, especially small objects. Therefore, we can make full use of these features in the top-level decoding process to guide the restoration of object details and address the issue of object size changes, and propose a parallel multi-scale attention (PMA) module to introduce the guidance from the low-level features in a multi-scale and attention manner. The multi-scale strategy is used to combat the varying object scale, and the attention strategy is to reduce inevitable redundant information in encoder.

\begin{figure}[t]
	\centering
	\includegraphics[width=\columnwidth]{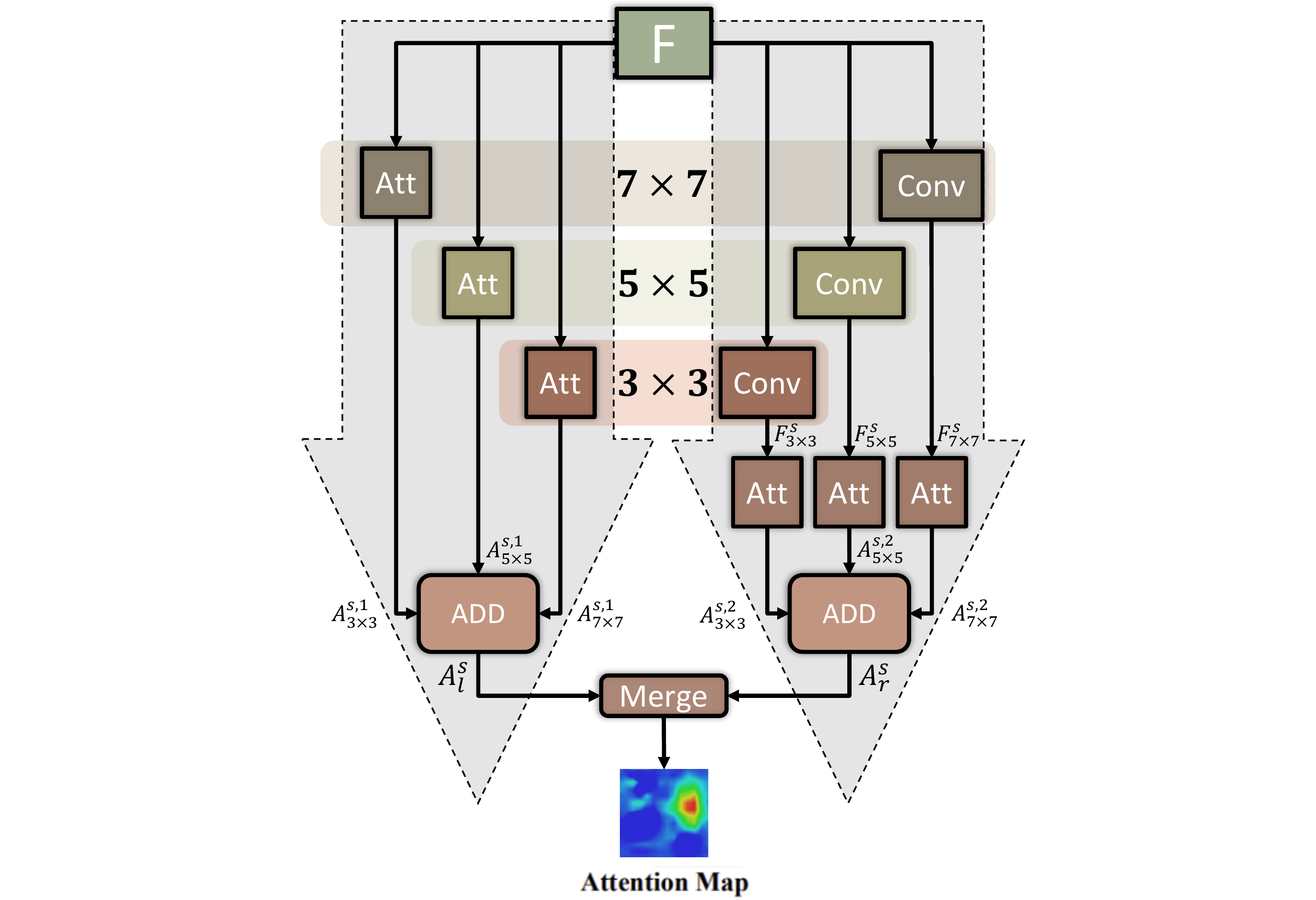}
	\caption { Illustration of parallel multi-scale attention (PMA) module, where ``Att'' refers to the spatial attention module, and ``Conv'' represents convolution operation with different kernel sizes.
	}
	\label{fig5}
\end{figure}

Different from the existing methods, we perform multi-scale calculation in the dimension of the attention map, and design two complementary attention calculation methods from the perspective of multi-scale feature generation. One is to directly perform multi-scale attention calculations on input features under different receptive fields to obtain attention maps at different scales, and then fuse them to generate the global multi-scale attention map, as shown in the left part of Fig. \ref{fig5}. The other is to first perform multi-scale feature extraction to obtain multi-scale features, and then perform attention calculation on each scale features, as shown in the right part of Fig. \ref{fig5}. The former emphasizes different attention information under different receptive fields, while the latter emphasizes the attention information of multi-scale features. In a word, we use the low-level convolutional features $X^s\in{\mathbb{R}^{C \times H \times W}}$ to calculate two multi-scale spatial attention maps $A_l^s$ and $A_r^s$, where $s\in{\{1,2\}}$ indexes the convolution stage. Finally, these two multi-scale attention maps are integrated into the final attention map to refine the up-sampled decoder features.

\subsubsection{Multi-scale attention on the single-scale features}For the spatial attention \cite{cbam}, the average-pooling and max-pooling are first applied on the input features, and then the convolutional layer and sigmoid activation are used to generate the final attention map. We replace the convolutional layer in the above process with multi-scale convolution to directly extract the multi-scale attention map of the input features. Specifically, we utilize the average-pooling and max-pooling on $X^s$ respectively to form two one-channel maps and then concatenate them along the channel axis, thereby generating a two-channel descriptor $\Gamma^s\in{\mathbb{R}^{ H \times W\times 2}}$:
\begin{equation}
	\Gamma^s=concat(avepool(X^s),maxpool(X^s)),
\end{equation}
where $concat(\cdot)$  represents feature concatenation along channel axis, $avepool(\cdot)$ and $maxpool(\cdot)$ are the average-pooling and max-pooling, respectively. The obtained two-channel descriptor is shown to be effective in highlighting informative regions. Then, convolution layers with different filter sizes of $3 \times 3$, $5 \times 5$, and $7 \times 7$ are conducted to turn a two-channel descriptor into three 2D spatial attention maps $A_{3 \times 3}^{s,1}\in{\mathbb{R}^{H \times W}}$, $A_{5\times5}^{s,1}\in{\mathbb{R}^{H \times W}}$, and $A_{7 \times 7}^{s,1}\in{\mathbb{R}^{H \times W}}$, which encode where to emphasize or suppress:
\begin{equation}
	\begin{array}{cc}
		& A_{3 \times 3}^{s,l}=\sigma(conv_{3\times3}(\Gamma^s;\hat{\theta}_{3\times3})), \\
		& A_{5 \times 5}^{s,l}=\sigma(conv_{5\times5}(\Gamma^s;\hat{\theta}_{5\times5})), \\
		& A_{7 \times 7}^{s,l}=\sigma(conv_{7\times7}(\Gamma^s;\hat{\theta}_{7\times7})),
	\end{array}
\end{equation}
where $\sigma$ denotes the sigmoid function, $conv_{n \times n}$ represents a convolution operation with the filter size of $n \times n$, and $\hat{\theta}_{n \times n}$ is the learnable parameters of the corresponding convolution operation.

Finally, these three attention maps with different receptive fields are aggregated to produce the final multi-scale attention map $A_l^s$, which is computed as:
\begin{equation}
	A_l^s=\frac{1}{3}(A_{3\times3}^{s,l} \oplus A_{5\times5}^{s,l} \oplus A_{7\times7}^{s,l}),
\end{equation}
where $\oplus$ represents the element-wise summation.

\subsubsection{Multi-scale attention on the multi-scale features}In this stream, we emphasize that the multi-scale attention is generated from the multi-scale features. In other words, different from the left stream, three convolution layers with the filter size of $3\times3$, $5\times5$, and $7\times7$ are separately applied on the input feature $X^s$, thereby generating multi-scale features $F_{3\times3}^s\in{\mathbb{R}^{H \times W \times C}}$,
$F_{5\times5}^s\in{\mathbb{R}^{ H \times W \times C }}$, and
$F_{7\times7}^s\in{\mathbb{R}^{H \times W \times C}}$:
\begin{equation}
	\begin{array}{cc}
		& F_{3\times3}^s=\sigma(conv_{3\times3}(X^s;\hat{\omega}_{3\times3})), \\
		& F_{5\times5}^s=\sigma(conv_{5\times5}(X^s;\hat{\omega}_{5\times5})), \\
		& F_{7\times7}^s=\sigma(conv_{7\times7}(X^s;\hat{\omega}_{7\times7})),
	\end{array}
\end{equation}
where $conv_{n \times n}$ represents a convolution operation with the filter size of $n \times n$. Then, we employ the spatial attention approach to these three multi-scale features, and generate three spatial attention maps
$A_{3 \times 3}^{s,r}\in{\mathbb{R}^{H \times W}}$,
$A_{5 \times 5}^{s,r}\in{\mathbb{R}^{H \times W}}$,
and $A_{7 \times 7}^{s,r}\in{\mathbb{R}^{H \times W}}$.
Analogy to the left stream, we aggregate three maps by addition to generate the multi-scale attention map $A_r^s$:
\begin{equation}
	A_r^s=\frac{1}{3}(A_{3\times3}^{s,r} \oplus A_{5\times5}^{s,r} \oplus A_{7\times7}^{s,r}).
\end{equation}
Finally, the multi-scale attentions generated from the single-scale and multi-scale features are combined into the final attention map
$A_f^s\in{\mathbb{R}^{H \times W}}$:
\begin{equation}
	A_f^s=\sigma(conv_{1\times1}(concat(A_l^s,A_r^s))).
\end{equation}
As a whole, we fuse passed-up deeper features $F_d^s$ with the corresponding shallower feature by concatenation and convolution, producing the fused features $F_d^{s-1}$, as shown in the upper right part of Fig. \ref{fig2}. In order to better restore the multi-scale details, the multi-scale attention generated from the shallower features is used to highlight the deep features. The fused features will be regarded as deep features to merge with the shallow features. Formally, the above fusion process can be denoted as:
\begin{flalign*}
	& F_d^{s-1}= &
\end{flalign*}
\begin{equation}
	\left\{
	\begin{array}{lr}
		\displaystyle conv(concat((F_d^s\uparrow)\odot(A_f^{s-1}+1),F_e^{s-1}))&s\in{\{2,3\}}\\
		\displaystyle conv(concat(F_d^s\uparrow,F_e^{s-1}))&s\in{\{4,5\}}
	\end{array},
	\right.
\end{equation}
where $\odot$ denotes element-wise multiplication with channel-wise broadcasting, $\uparrow$ means $2\times$ spatial up-sampling operation. $F_e^{s-1}$ is the corresponding encoder features, when $s\in{\{2,3\}}$, $F_e^{s-1}=X^{s-1}$, otherwise, $F_e^{s-1}=F_{rc}^{s-1}$.
\begin{figure*}[!t]
	\centering
	\includegraphics[width=1\linewidth]{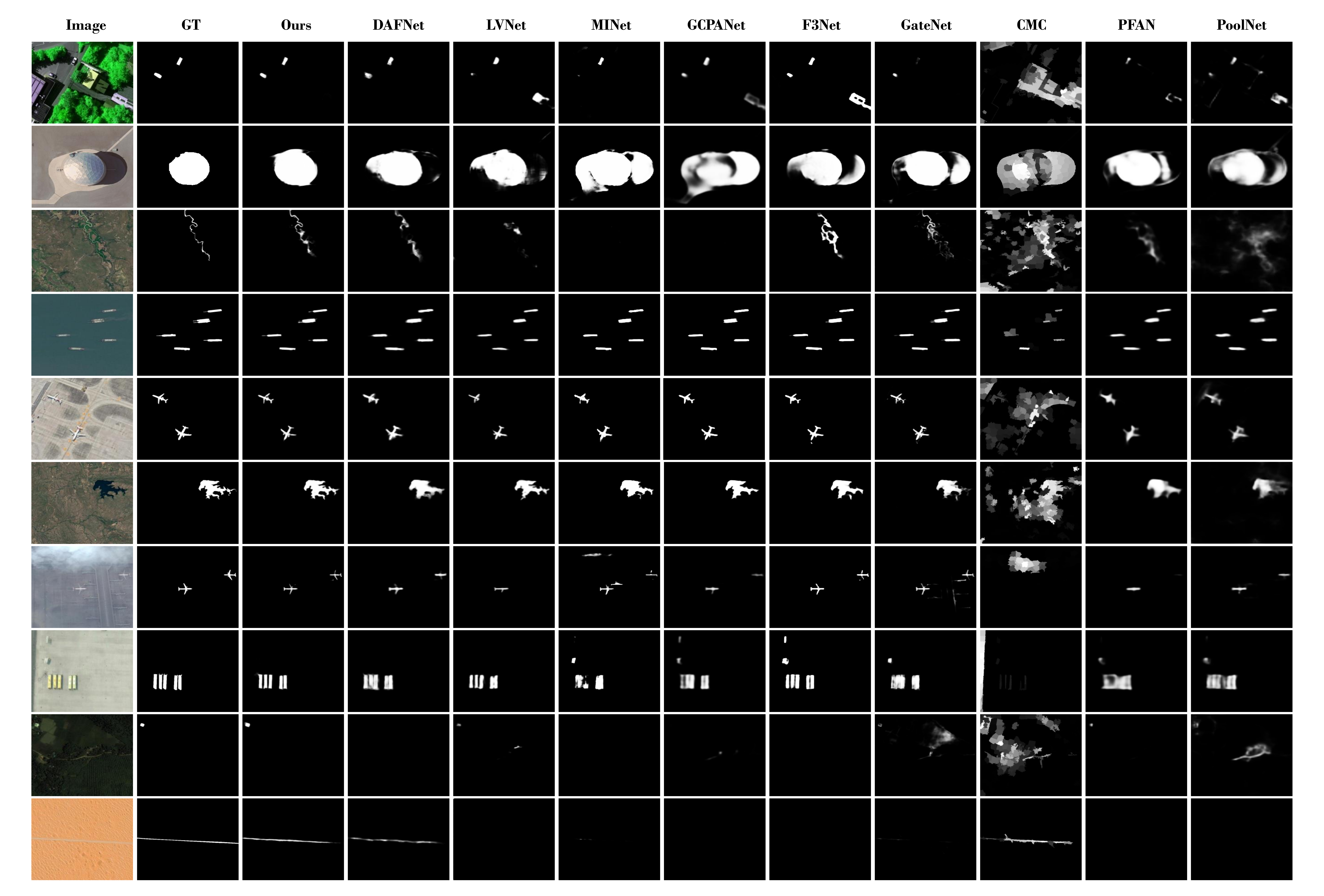}
	\caption{Visual comparisons of our proposed method and SOTA methods on EORSSD dataset. The deep learning based methods are trained/re-trained on the EORSSD dataset.
	}
	\label{fig6}
\end{figure*}

\subsection{Loss Function}
The final saliency map of our network are derived from features at the top decoding level. We use the class-balanced binary cross-entropy as the loss function for saliency prediction supervision:
\begin{equation}
	\ell=-[p\cdot\mathcal{L}log(\mathcal{S})+q\cdot(1-\mathcal{L})log(1-\mathcal{S})],
\end{equation}
where $\mathcal{S}$ is the predicted saliency map, $\mathcal{L}$ is the saliency ground truth label, $p=(\mathcal{B}-\mathcal{B}_m)/\mathcal{B}$ and $q=\mathcal{B}_m/\mathcal{B}$ are used to balance the contribution of salient and background pixels. $\mathcal{B}$ denotes the number of pixels in image and $\mathcal{B}_m$ represents the number of positive numbers in label $\mathcal{L}$.

\section{Experiments} \label{sec4}

\subsection{Evaluation Metrics and Implementation Details}
\textbf{Evaluation Metrics.} We employ precision-recall (P-R) curve, F-measure ($F_\beta$), Mean Absolute Error (MAE), E-measure ($E_m$), and S-measure ($S_m$) to quantitatively evaluate our proposed model performance \cite{d3net}. The P-R curve being closer to the upper right means better performance. For other metrics, the larger the F-measure, the S-measure, and the E-measure, stands for the better performance, where as the smaller MAE score means the better the performance. In this paper, we uniformly adopted the widely used evaluation code \cite{d3net} based on Matlab (Download link: http://dpfan.net/d3netbenchmark/).

\textbf{Implementation Details.} We train and test on the ORSSD \cite{LVNet} and EORSSD \cite{DAFNet} databases, respectively. The ORSSD dataset includes $600$ images for training and $200$ images for testing, and the EORSSD dataset contains $1,400$ training images and $600$ testing images. Data augmentation techniques involving combinations of flipping and rotation are employed to improve the diversity of training samples, which produces seven variants for every single sample. Moreover, each sample is uniformly resized to $224\times224$ due to the limited computing resources.
Following the settings in \cite{DAFNet}, the Res2Net-50 \cite{Res2Net} pretrained on ImageNet \cite{imagenet} is employed as the backbone feature extractor in the experiment. The proposed RRNet is implemented with PyTorch and deployed on a workstation equipped with an NVIDIA GeForce RTX 2080Ti GPU. And we also implement our network by using the MindSpore Lite tool\footnote{https://www.mindspore.cn/}. The ADAM optimization strategy is utilized to train model parameters for $35,000$ iterations, and the batch size is set to $8$. The learning rate is set to $5e^{-5}$ for the initial learning rate, and then evenly decrease to $5e^{-7}$. The method of initializing weights we use is Xavier policy \cite{Xavier}, and the bias parameters were initialized as constants. The proposed model has a real-time inference speed of $109$ FPS for processing an image with the size of $224\times224$.

\subsection{Comparison with State-of-the-art Methods}

In experiment, we compare the proposed RRNet with thirteen SOD methods, including four SOD methods for optical RSIs (\ie, DAFNet \cite{DAFNet}, LVNet \cite{LVNet}, CMC \cite{CMC}, and SMFF \cite{SMFF}), nine SOD methods for NSIs published in 2020 and 2019 (\ie, MINet \cite{MINet}, GCPANet \cite{GCPANet}, F3Net \cite{F3Net}, GateNet \cite{GateNet}, PoolNet \cite{PoolNet}, PFAN \cite{PFAN}, EGNet \cite{EGNet}, RADF \cite{RADF}, and R3Net \cite{R3Net}). We directly use either the source codes or the saliency maps provided by the authors on the testing subset of ORSSD dataset and EORSSD dataset. For a fair comparison, we retain SOD methods for NSIs to generate the results by implementing with the recommended parameter settings.

\subsubsection{\textbf{Qualitative Comparison}}
Some visual comparisons are shown in Fig. \ref{fig6}. It can be seen that our model can more accurately and completely detect salient objects in complex scenes by using global semantic relations and multi-scale detail restoration. Our advantages are expressed in the following aspects:
\textbf{(a) Advantages in cluttered background.} Thanks to the relational reasoning module, our model can suppress the cluttered background interference and highlight the interior of foreground regions. For example, in the second image containing both light-colored interference and dark shadow, other SOD methods for NSIs and RSIs fail to suppress them together. Similarly, in the third image, other methods cannot locate the long-range and wandering river accurately and completely. By contrast, our model achieves better performance in location accuracy and background suppression.
\textbf{(b) Advantages in detail information.} Our model has a superior ability to detect detail information such as boundaries and structures. For example, in the fourth and fifth images, the detail structures (\eg, bowsprits of ships and aero-engines of planes) are lost by other methods. In the sixth image, the shorelines of the lake detected by all other methods are ambiguous and incomplete. Turning to our method, the boundaries are more accurate than others, and the detailed structures are more complete.
\textbf{(c) Advantages in multi-object detection.} For the multi-objects scene, our model accurately detects all the salient objects by reasoning the relationship between them. For example, in the seventh image, the lower-contrast airplane on the upper right caused by the cloud is difficult to locate accurately, but our method can still detect it more accurately and completely. In the eighth image,  compared with other methods, our method can not only detect all salient targets, but also maintain a clear position boundary between different objects like the ground truth.
\textbf{(d) Other challenging scenarios.} By taking advantage of different proposed modules, our model can handle various complex scenarios. In the ninth image, it proves that our model is not sensitive to the location and size of the salient object, and it can still work well even when the object is placed far away from the image center or is very small. In addition, in the tenth image, the similar color of road and desert increases the difficulty of detection, but our method can still handle this low-contrast case and generate relatively complete salient regions.
\begin{figure}[t]
	\centering
	\includegraphics[width=\columnwidth]{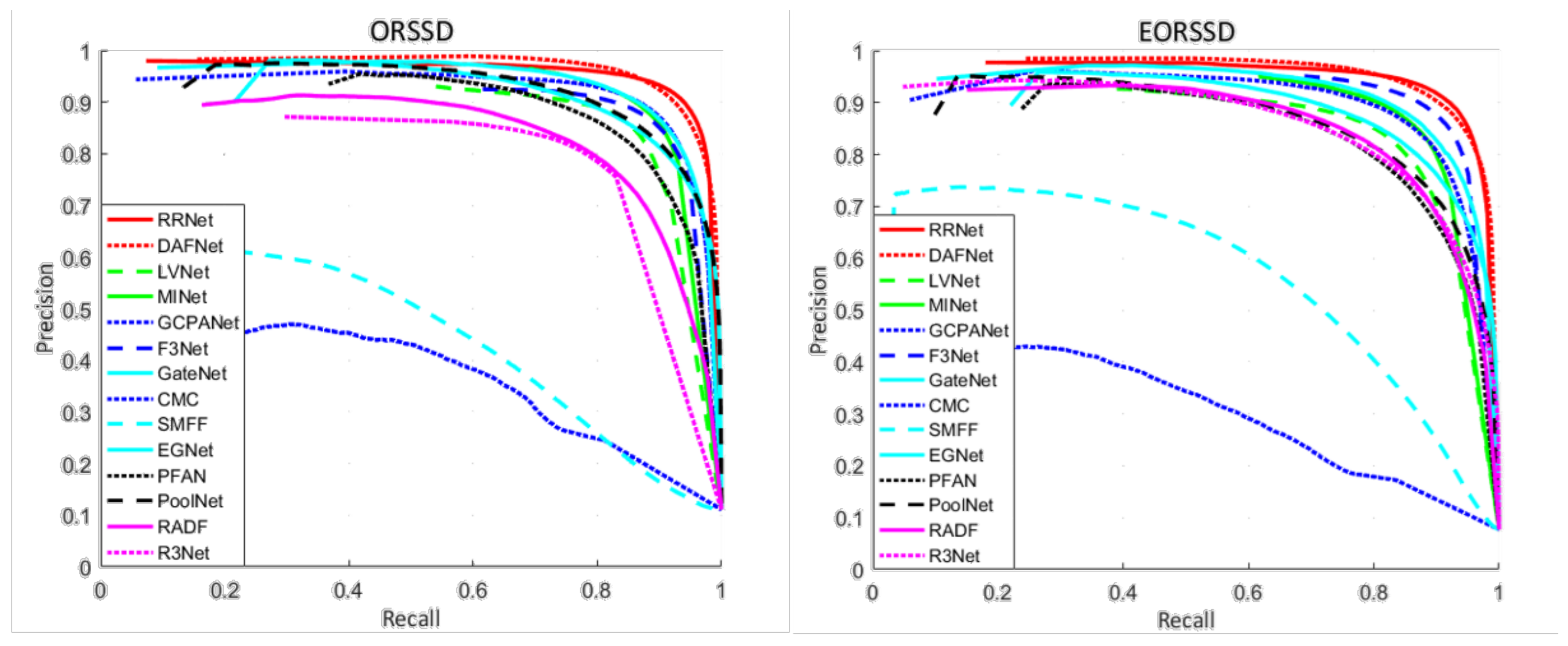}
	\caption {The PR curves of different methods on the ORSSD dataset and EORSSD dataset respectively.
	}
	\label{fig7}
\end{figure}
\begin{table}[!t]
	\scriptsize
	\renewcommand{\arraystretch}{1.3}
	\label{table_example}
	
	\caption{Quantitative Comparisons with Different Methods on the Testing Subset of ORSSD and EORSSD Datasets. The Best Three Results Are Marked in Bold.}
	\setlength{\tabcolsep}{1.5mm}{
	\centering
	\begin{tabular}{c|cccc|cccc}
		\hline\hline
		& \multicolumn{4}{c|}{ORSSD Dataset}                                                                                                           & \multicolumn{4}{c}{EORSSD Dataset}                                                                                                          \\ \cline{2-9}
		& $F_\beta$                                 & $E_m$      & MAE                                       & $S_m$                                     & $F_\beta$                                 & $E_m$      & MAE                                       & $S_m$                                     \\ \hline\hline
		R3Net   & $.7698$                                  & $.8907$ & $.0409$                                  & $.8092$                                  & $.7989$                                  & $.9547$ & $.0170$                                  & $.8305$                                  \\ \hline
		RADF    & $.7865$                                  & $.9123$ & $.0386$                                  & $.8252$                                  & $.7966$                                  & $.9227$ & $.0162$                                  & $.8332$                                  \\ \hline
		PoolNet & $.7911$                                  & $.9604$ & $.0358$                                  & $.8403$                                  & $.8012$                                  & $.9358$ & $.0209$                                  & $.8301$                                  \\ \hline
		PFAN    & $.8344$                                  & $.9418$ & $.0543$                                  & $.8613$                                  & $.7931$                                  & $.9334$ & $.0156$                                  & $.8446$                                  \\ \hline
		EGNet   & $.8585$                                  & $.9727$ & $.0215$                                  & $.8780$                                  & $.8310$                                  & $.9600$ & $.0109$                                  & $.8692$                                  \\ \hline
		GateNet & $.8794$                                  & $.9464$ & $.0197$                                  & $.8853$                                  & $.8618$                                & $.9440$ & $.0131$                                  & $.8710$                                  \\ \hline
		F3Net   & $.8661$                                  & $.9433$ & $.0215$                                  & $.8949$ & $.8681$ & $.9487$ & $.0119$                                  & $.9040$ \\ \hline
		GCPANet & $.8833$ & $.9545$ & $.0186$                                  & $.8865$                                  & $.8546$                                  & $.9448$ & $.0123$                                  & $.8674$                                  \\ \hline
		MINet   & $.8751$                                  & $.9423$ & $.0171$ & $.8865$                                  & $.8510$                                  & $.9354$ & $.0104$ & $.8909$                                  \\ \hline
		SMFF    & $.4764$                                  & $.7518$ & $.1897$                                  & $.5329$                                  & $.5693$                                  & $.7892$ & $.1471$                                  & $.5431$                                  \\ \hline
		CMC     & $.4214$                                  & $.7069$ & $.1267$                                  & $.6033$                                  & $.3555$                                  & $.6785$ & $.1066$                                  & $.5826$                                  \\ \hline
		LVNet   & $.8414$                                  & $.9342$ & $.0207$                                  & $.8815$                                  & $.8213$                                  & $.9302$ & $.0146$                                  & $.8642$                                  \\ \hline
		DAFNet  & $.9192$ & $.9699$ & $.0105$ & $.9188$ & $.9060$ & $.9684$ & $\textbf{.0053}$ & $.9185$ \\ \hline
		Ours    & $\textbf{.9203}$ & $\textbf{.9808}$ & $\textbf{.0103}$ & $\textbf{.9282}$ & $\textbf{.9119}$ & $\textbf{.9720}$ & $.0076$ & $\textbf{.9230}$ \\ \hline\hline
	\end{tabular}}
\label{tab1}
\end{table}
\subsubsection{\textbf{Quantitative Comparison}}
The P-R curves of all methods on the ORSSD dataset and EORSSD dataset are shown in Fig. \ref{fig7}. As visible, the proposed RRNet has a higher position on the P-R curve than other methods, convincingly demonstrating the effectiveness of our model.
Besides, the numerical indexes including the S-measure ($S_m$), E-measure ($E_m$), MAE score, and F-measure ($F_\beta$) are reported in Table \ref{tab1}. It is evident that our model achieves competitive performance on these two datasets across different metrics. To be specific, in addition to the MAE score on the EORSSD dataset, our method achieves the best performance in terms of all measures on the two datasets. Compared with the non-deep learning based SOD methods for optical RSIs (\ie, SMFF \cite{SMFF} and CMC \cite{CMC}), all the deep learning based methods including the models for NSIs exhibit the superior performance. Moreover, we can see that the SOD methods specifically designed for optical RSIs (\ie, DAFNet \cite{DAFNet}, LVNet \cite{LVNet}, and ours) generally achieve better performance than the retrained SOD methods for NSIs, demonstrating the particularity and challenges of SOD in optical RSIs.
In summary, thanks to the delicately designed modules, our RRNet ranks first in terms of the S-measure, E-measure, and F-measure. For example, compared with the \textbf{second best} method, the percentage gain reaches $0.8\%$ in terms of the E-measure, $1.9\%$ in terms of the MAE score, and $1.0\%$ in terms of S-measure on the ORSSD dataset. On the EORSSD dataset, the \textbf{minimum percentage gain} reaches $0.5\%$ in terms of the S-measure and $0.7\%$ in terms of the F-measure. All these clearly demonstrate the superior performance of proposed model in SOD for optical RSIs.


\begin{figure}[t]
	\centering
	\includegraphics[width=1\columnwidth]{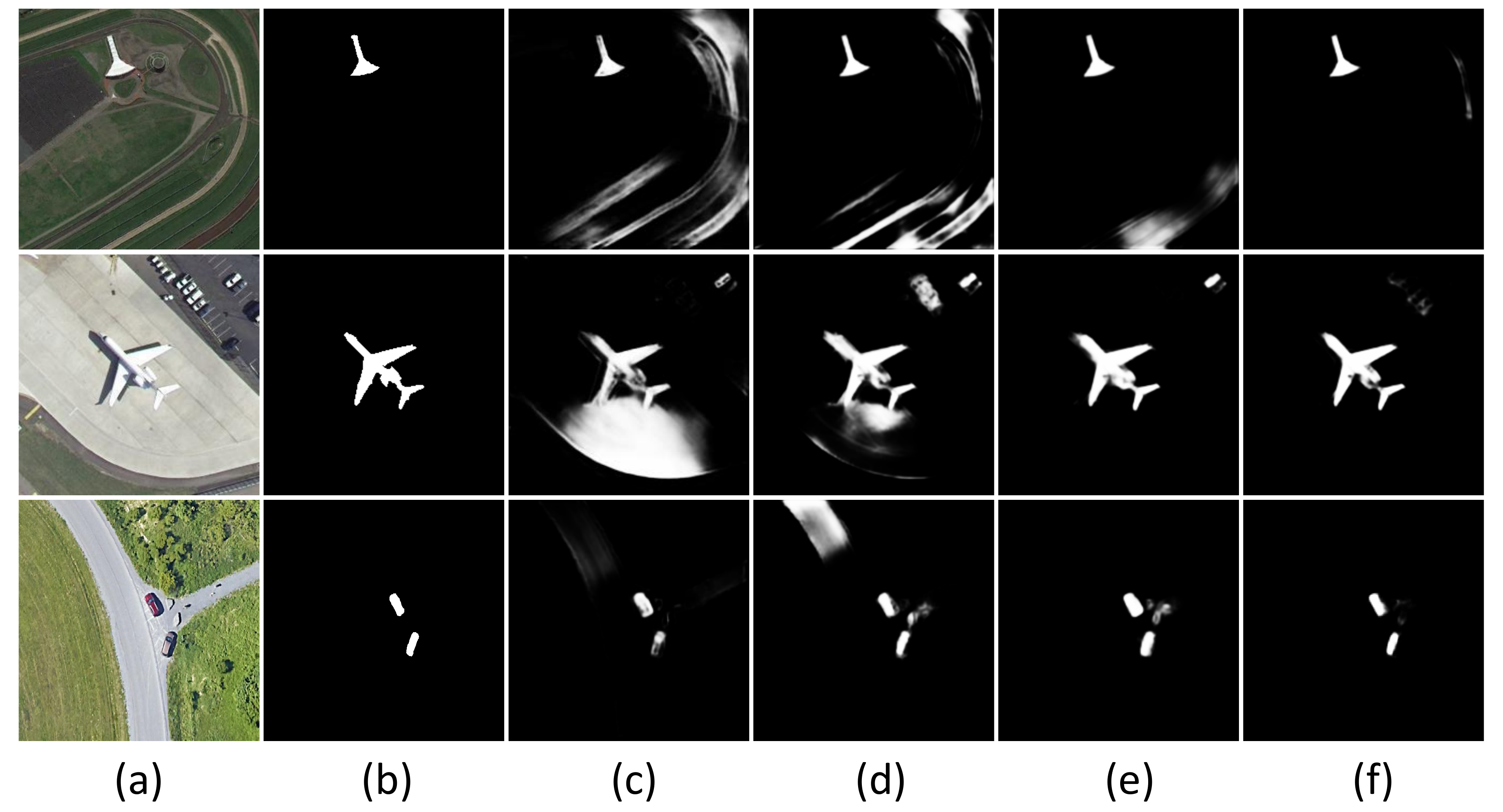}
	\caption {Visual Comparison of RRNet variants equipped with different models. (a) Optical RSIs. (b) Ground truth. (c) Baseline. (d) Baseline + PMA. (e) Baseline + PMA + SRR. (f) Baseline + PMA + SRR + CRR.
	}
	\label{fig8}
\end{figure}

\subsection{Ablation Study}
We conduct ablation experiments to demonstrate the effectiveness of each component in our model, mainly including the parallel multi-scale attention (PMA) module and relational reasoning module. Furthermore, we split relational reasoning module into spatial relational reasoning and channel relational reasoning. We present the qualitative comparison results in Fig. \ref{fig8}. Compared with the baseline (\ie, Res2Net-50 \cite{Res2Net}), it can be found that PMA model can capture more detail information and obtain clearer boundaries (\eg, the plane in the second image and the cars in the third image), but some interferences are also detected. Applying the spatial relational reasoning, as shown in Fig. \ref{fig8}(d), most of the interferences can be suppressed by reasoning the context information (\eg, the car and runway in the second image, and the roads in the first image and the third image). However, spatial relational reasoning only constructs a graph of spatial locations, leading to the underuse of features and incompletely suppression of background information. By contrast, the proposed RRNet which contains spatial relational reasoning and channel relational reasoning can accurately locate the salient objects and fully suppress the backgrounds (\eg, the road in the first image and the stones in the third image).

\begin{table}[!t]
	\renewcommand{\arraystretch}{1.3}
	\caption{Ablation Analysis on the EORSSD Dataset.}
	\label{table_example}
	\setlength{\tabcolsep}{1.8mm}{
	\centering
	\begin{tabular}{cccc|cccc}
		\hline \hline
		Baseline   & PMA        & SRR        & CRR        & $F_\beta$   & $E_m$    & MAE    & $S_m$  \\ \hline
		
		\checkmark &            &            &            & $0.8302$    & $0.9217$ & $0.0148$ & $0.8695$ \\
		
		\checkmark & \checkmark &            &            & $0.8819$    & $0.9523$ & $0.0105$ & $0.9021$ \\
		
		\checkmark & \checkmark & \checkmark &            & $0.8947$    & $0.9582$ & $0.0091$ & $0.9156$ \\
		
		\checkmark & \checkmark & \checkmark & \checkmark & $\textbf{0.9119}$    & $\textbf{0.9720}$ & $\textbf{0.0076}$ & $\textbf{0.9230}$  \\ \hline \hline
	\end{tabular}}
	\label{tab2}
\end{table}

\begin{table}[!t]

	\renewcommand{\arraystretch}{1.3}
	\caption{Further validation of RR and PMA on the EORSSD Dataset.}
	\label{table_example}
	\setlength{\tabcolsep}{2.6mm}{
		\centering
		\begin{tabular}{c|c|c|c|c|c}
			\hline\hline
			\multicolumn{2}{c|}{Modules}             & $F_\beta$         & $E_m$             & MAE               & $S_m$             \\ \hline
			\multicolumn{2}{c|}{\textbf{full model}} & $\textbf{0.9119}$ & $\textbf{0.9720}$ & $\textbf{0.0076}$ & $\textbf{0.9230}$ \\ \hline
RR                      & w/Non-local    & {$0.9102$}        & {$0.9691$}        & {$0.0093$}        & {$0.9225$}        \\\hline
			\multirow{2}{*}{PMA}    & w/o PMA(r)     & {$0.9100$}        & {$0.9707$}        & {$0.0079$}        & {$0.9227$}        \\
			& w/o PMA(l)     & {$0.9037$}        & {$0.9544$}        & {$0.0089$}        & {$0.9094$}        \\ \hline\hline
		\end{tabular}}
	\label{tab3}
\end{table}

We provide the quantitative results of variants equipped with different modules in Table \ref{tab2}. The PMA module is first added to baseline to enhance the extraction of detail information and address the scale variation of objects. As seen from this table, the F-measure is improved from $0.8302$ to $0.8819$ with a percentage gain of $3.7\%$, and the S-measure is boosted from $0.8695$ to $0.9021$ with the percentage gain of $6.2\%$. Then, we add the spatial relational reasoning to the aforementioned architecture to deal with the complex scenarios. From Table \ref{tab2}, it can be observed that the use of the spatial relational reasoning module further improves the saliency detection performance with the percentage gain of $13.3\%$ in terms of MAE score. Finally, the full model with the channel relational reasoning achieves the best performance.

In order to further verify the superiority of the proposed relational reasoning module, we replace RR module with Non-local module and remains other modules unchanged. The quantitative comparisons are shown in Table \ref{tab3}. We can see that RR module is more effective than the Non-local module in SOD task. For example, the MAE score is turned from $0.0076$ to $0.0093$ with a percentage drop of $22.3\%$ when the model equipped with Non-local module (\ie, w/ Non-local). In addition, we provide the quantitative results of variants equipped with only the left or right branch respectively in Table \ref{tab3}. We can clearly see that the absence of any branch in the PMA module leads to inferior performance. When only the left branch is retained (\ie, w/o PMA(r)), the F-measure is declined from $0.9119$ to $0.9100$ with a percentage drop of $0.2\%$. The F-measure is down by $0.9\%$ when only the right branch is retained (\ie, w/o PMA(l)). All the visual examples and quantitative results demonstrate the effectiveness of the key modules designed in our RRNet.

\begin{figure}[t]
	\centering
	\includegraphics[width=1\linewidth,height=0.4\linewidth]{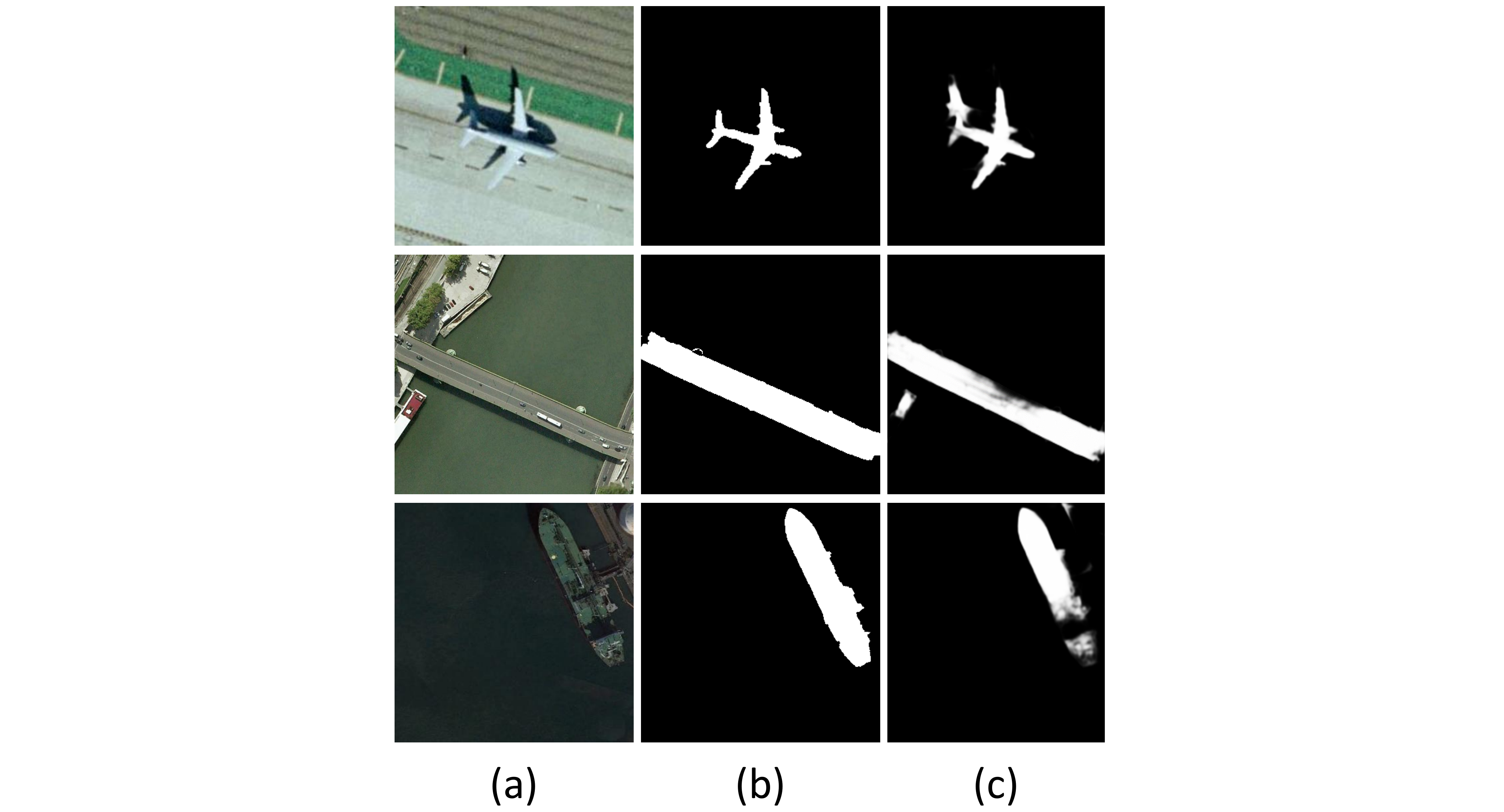}
	\caption {Visual examples of failure cases. (a) Optical RSIs. (b) Ground truth. (c) Saliency maps produced by RRNet.}
	\label{fig9}
\end{figure}

\subsection{Discussion}
\subsubsection{\textbf{Failure Cases}}
Some failure cases are shown in Fig. \ref{fig9}. For some very challenging examples, our method still cannot obtain perfect results.

\textbf{(a) It is still challenging to deal with complex shadows for our method.} For example, in the first row of Fig. \ref{fig9}, the shadows on the fuselage can be suppressed, but the shadows on the rear of the aircraft are wrongly reserved.
\textbf{(b) It is still challenging to well suppress the high-contrast but non-salient objects.} For example, in the second row of Fig. \ref{fig9}, the red box in the lower left corner contrasts sharply with the white boat. But apparently, the bridge is the salient object. Our method wrongly detects them as the salient objects due to the strong interference from bright red attribute.
\textbf{(c) It is still challenging to completely detect salient object.} For example, in the third row of Fig. \ref{fig9}, the stern of the ship cannot be completely detected due to its high appearance similarity with surrounding contexts.

\subsubsection{\textbf{Future Work}}
First, the MAE of our method only achieves the second best performance against the competitors. For the challenging examples mentioned in failure cases, we can make some attempts to optimize them. For example, for the case where the appearance of salient object is very close to the background, some techniques for camouflaged object detection (COD) task \cite{COD-1,COD-2} can be used for reference, such as texture exploration.
Second, in addition to RGB channels, other spectral coverage (such as a near-infrared channels which are invisible to humans) can provide more comprehensive and richer information, which may help further boost the detection accuracy. In the future, we would like to construct a large-scale multispectral benchmark dataset for SOD in order to further improve the prediction ability of our model.

\section{Conclusion} \label{sec5}
In this paper, a novel end-to-end SOD model for optical RSIs is presented, named RRNet, which is capable of reasoning semantic information and restoring detail information. The relational reasoning in the spatial space and channel space is designed to model the relationship between different salient objects or different parts of the salient object, which can effectively suppress background interference and force the generation of complete salient objects. Moreover, we propose a parallel multi-scale attention module that utilizes attention mechanism to improve the detection accuracy and restore the details of different scale objects through multi-scale design. Experimental evaluations over two datasets indicate that our method outperforms the state-of-the-art salient object detectors. Furthermore, when this work is extended for remote sensing images, our relational reasoning module can be used as a plug-and-play module to model the semantic relation of different objects or different parts of objects on the graph, thereby enhancing performance.

\par
\ifCLASSOPTIONcaptionsoff
  \newpage
\fi
{
\bibliographystyle{IEEEtran}
\bibliography{ref}

\begin{thebibliography}{10}
\providecommand{\url}[1]{#1}
\csname url@samestyle\endcsname
\providecommand{\newblock}{\relax}
\providecommand{\bibinfo}[2]{#2}
\providecommand{\BIBentrySTDinterwordspacing}{\spaceskip=0pt\relax}
\providecommand{\BIBentryALTinterwordstretchfactor}{4}
\providecommand{\BIBentryALTinterwordspacing}{\spaceskip=\fontdimen2\font plus
\BIBentryALTinterwordstretchfactor\fontdimen3\font minus
  \fontdimen4\font\relax}
\providecommand{\BIBforeignlanguage}[2]{{%
\expandafter\ifx\csname l@#1\endcsname\relax
\typeout{** WARNING: IEEEtran.bst: No hyphenation pattern has been}%
\typeout{** loaded for the language `#1'. Using the pattern for}%
\typeout{** the default language instead.}%
\else
\language=\csname l@#1\endcsname
\fi
#2}}
\providecommand{\BIBdecl}{\relax}
\BIBdecl

\bibitem{crm_review}
R.~Cong, J.~Lei, H.~Fu, M.-M. Cheng, W.~Lin, and Q.~Huang, ``Review of visual
  saliency detectioin with comprehensive information,'' \emph{IEEE Trans.
  Circuits Syst. Video Technol}, vol.~29, no.~10, pp. 2941--2959, 2019.

\bibitem{intro-3}
W.~Wang, J.~Shen, R.~Yang, and F.~Porikli, ``Saliency-aware video object
  segmentation,'' \emph{IEEE Trans. Pattern Anal. Mach. Intell.}, vol.~40,
  no.~1, pp. 20--33, 2018.

\bibitem{intro-1}
Y.~Fang, Z.~Chen, W.~Lin, and C.-W. Lin, ``Saliency detection in the compressed
  domain for adaptive image retargeting,'' \emph{IEEE Trans. Image Process.},
  vol.~21, no.~9, pp. 3888--3901, 2012.

\bibitem{intro-4}
W.~Wang, J.~Shen, Y.~Yu, and K.-L. Ma, ``Stereoscopic thumbnail creation via
  efficient stereo saliency detection,'' \emph{IEEE Trans. Vis. Comput. Graph},
  vol.~23, no.~8, pp. 2014--2027, 2017.

\bibitem{intro-6}
K.~Gu, S.~Wang, H.~Yang, W.~Lin, G.~Zhai, X.~Yang, and W.~Zhang,
  ``Saliency-guided quality assessment of screen content images,'' \emph{IEEE
  Trans. Multimedia}, vol.~18, no.~6, pp. 1098--1110, 2016.

\bibitem{yuan2019vssa}
Y.~Yuan, Z.~Xiong, and Q.~Wang, ``{VSSA-NET}: vertical spatial sequence
  attention network for traffic sign detection,'' \emph{IEEE Trans. Image
  Process.}, vol.~28, no.~7, pp. 3423--3434, 2019.

\bibitem{wang2020multitask}
Q.~Wang, T.~Han, Z.~Qin, J.~Gao, and X.~Li, ``Multitask attention network for
  lane detection and fitting,'' \emph{IEEE Trans. Neural Netw. Learn. Syst.},
  pp. 1--13, 2020.

\bibitem{li2017multiview}
X.~Li, M.~Chen, F.~Nie, and Q.~Wang, ``A multiview-based parameter free
  framework for group detection,'' in \emph{Proc. AAAI}, 2017, pp. 4147--4153.

\bibitem{R3Net}
Z.~Deng, X.~Hu, L.~Zhu, X.~Xu, J.~Qin, G.~Han, and P.-A. Heng, ``R$^{3}${N}et:
  Recurrent residual refinement network for saliency detection,'' in
  \emph{Proc. IJCAI}, 2018, pp. 684--690.

\bibitem{RADF}
X.~Hu, L.~Zhu, J.~Qin, C.~W. Fu, and P.~A. Heng, ``Recurrently aggregating deep
  features for salient object detection,'' in \emph{Proc. AAAI}, 2018, pp.
  6943--6950.

\bibitem{PFAN}
T.~Zhao and X.~Wu, ``Pyramid feature attention network for saliency
  detection,'' in \emph{Proc. CVPR}, 2019, pp. 3085--3094.

\bibitem{MINet}
Y.~Pang, X.~Zhao, L.~Zhang, and H.~Lu, ``Multi-scale interactive network for
  salient object detection,'' in \emph{Proc. CVPR}, 2020, pp. 9410--9419.

\bibitem{GateNet}
X.~Zhao, Y.~Pang, L.~Zhang, H.~Lu, and L.~Zhang, ``Suppress and balance: {A}
  simple gated network for salient object detection,'' in \emph{Proc. ECCV},
  2020, pp. 35--51.

\bibitem{F3Net}
J.~Wei, S.~Wang, and Q.~Huang, ``F{\({^3}\)}net: Fusion, feedback and focus for
  salient object detection,'' in \emph{Proc. AAAI}, 2020, pp. 12\,321--12\,328.

\bibitem{EGNet}
J.-X. Zhao, J.-J. Liu, D.-P. Fan, Y.~Cao, J.-F. Yang, and M.-M. Cheng,
  ``{EGNet}: Edge guidance network for salient object detection,'' in
  \emph{Proc. ICCV}, 2019, pp. 8779--8788.

\bibitem{PoolNet}
J.-J. Liu, Q.~Hou, M.-M. Cheng, J.~Feng, and J.~Jiang, ``A simple pooling-based
  design for real-time salient object detection,'' in \emph{Proc. CVPR}, 2019,
  pp. 3917--3926.

\bibitem{GCPANet}
Z.~Chen, Q.~Xu, R.~Cong, and Q.~Huang, ``Global context-aware progressive
  aggregation network for salient object detection,'' in \emph{Proc. AAAI},
  2020, pp. 10\,599--10\,606.

\bibitem{RGBD-1}
C.~Li, R.~Cong, S.~Kwong, J.~Hou, H.~Fu, G.~Zhu, D.~Zhang, and Q.~Huang,
  ``{ASIF-Net}: Attention steered interweave fusion network for {RGBD} salient
  object detection,'' \emph{IEEE Trans. Cybern.}, vol.~51, no.~1, pp. 88--100,
  2020.

\bibitem{RGBD-3}
Z.~Chen, R.~Cong, Q.~Xu, and Q.~Huang, ``Dpanet: Depth potentiality-aware gated
  attention network for rgb-d salient object detection,'' \emph{IEEE Trans.
  Image Process.}, 2020.

\bibitem{crmspl}
R.~Cong, J.~Lei, C.~Zhang, Q.~Huang, X.~Cao, and C.~Hou, ``Saliency detection
  for stereoscopic images based on depth confidence analysis and multiple cues
  fusion,'' \emph{IEEE Signal Process. Lett.}, vol.~23, no.~6, pp. 819--823,
  2016.

\bibitem{crmtc20}
R.~Cong, J.~Lei, H.~Fu, J.~Hou, Q.~Huang, and S.~Kwong, ``Going from {RGB} to
  {RGBD} saliency: A depth-guided transformation model,'' \emph{IEEE Trans.
  Cybern.}, vol.~50, no.~8, pp. 3627--3639, 2020.

\bibitem{eccv20}
C.~Li, R.~Cong, Y.~Piao, Q.~Xu, and C.~C. Loy, ``{RGB-D} salient object
  detection with cross-modality modulation and selection,'' in \emph{Proc.
  ECCV}, 2020, pp. 225--241.

\bibitem{nips20}
Q.~Zhang, R.~Cong, J.~Hou, C.~Li, and Y.~Zhao, ``{CoADNet}: Collaborative
  aggregation-and-distribution networks for co-salient object detection,'' in
  \emph{Proc. NeurIPS}, 2020.

\bibitem{crmtmm19}
R.~Cong, J.~Lei, H.~Fu, Q.~Huang, X.~Cao, and N.~Ling, ``{HSCS}: Hierarchical
  sparsity based co-saliency detection for {RGBD} images,'' \emph{IEEE Trans.
  Multimedia}, vol.~21, no.~7, pp. 1660--1671, 2019.

\bibitem{crmtip18}
R.~Cong, J.~Lei, H.~Fu, Q.~Huang, X.~Cao, and C.~Hou, ``Co-saliency detection
  for {RGBD} images based on multi-constraint feature matching and cross label
  propagation,'' \emph{IEEE Trans. Image Process.}, vol.~27, no.~2, pp.
  568--579, 2018.

\bibitem{crmtc19}
R.~Cong, J.~Lei, H.~Fu, W.~Lin, Q.~Huang, X.~Cao, and C.~Hou, ``An iterative
  co-saliency framework for {RGBD} images,'' \emph{IEEE Trans. Cybern.},
  vol.~49, no.~1, pp. 233--246, 2019.

\bibitem{Video-1}
R.~Cong, J.~Lei, H.~Fu, F.~Porikli, Q.~Huang, and C.~Hou, ``Video saliency
  detection via sparsity-based reconstruction and propagation,'' \emph{IEEE
  Trans. Image Process.}, vol.~28, no.~10, pp. 4819--4831, 2019.

\bibitem{DAFNet}
Q.~Zhang, R.~Cong, C.~Li, M.-M. Cheng, Y.~Fang, X.~Cao, Y.~Zhao, and S.~Kwong,
  ``Dense attention fluid network for salient object detection in optical
  remote sensing images,'' \emph{IEEE Trans. Image Process.}, vol.~30, pp.
  1305--1317, 2021.

\bibitem{LVNet}
C.~Li, R.~Cong, J.~Hou, S.~Zhang, Y.~Qian, and S.~Kwong, ``Nested network with
  two-stream pyramid for salient object detection in optical remote sensing
  images,'' \emph{IEEE Trans. Geosci. Remote Sens.}, vol.~57, no.~11, pp.
  9156--9166, 2019.

\bibitem{lcync20}
C.~Li, R.~Cong, C.~Guo, H.~Li, C.~Zhang, F.~Zheng, and Y.~Zhao, ``A parallel
  down-up fusion network for salient object detection in optical remote sensing
  images,,'' \emph{Neurocomputing}, vol. 415, pp. 411--420, 2020.

\bibitem{SMFF}
L.~Zhang, Y.~Liu, and J.~Zhang, ``Saliency detection based on self-adaptive
  multiple feature fusion for remote sensing images,'' \emph{Int. J. Remote
  Sens.}, vol.~40, no.~22, pp. 8270--8297, 2019.

\bibitem{wu2019orsim}
X.~Wu, D.~Hong, J.~Tian, J.~Chanussot, W.~Li, and R.~Tao, ``{ORSIm} detector: A
  novel object detection framework in optical remote sensing imagery using
  spatial-frequency channel features,'' \emph{IEEE Trans. Geosci. Remote
  Sens.}, vol.~57, no.~7, pp. 5146--5158, 2019.

\bibitem{li2020object}
K.~Li, G.~Wan, G.~Cheng, L.~Meng, and J.~Han, ``Object detection in optical
  remote sensing images: A survey and a new benchmark,'' \emph{ISPRS J.
  Photogramm.}, vol. 159, pp. 296--307, 2020.

\bibitem{rsi6}
C.~Dong, J.~Liu, and F.~Xu, ``Ship detection in optical remote sensing images
  based on saliency and a rotation-invariant descriptor,'' \emph{Remote Sens.},
  vol.~10, no.~3, pp. 1--19, 2018.

\bibitem{yin2019region}
S.~Yin, Y.~Zhang, and S.~Karim, ``Region search based on hybrid convolutional
  neural network in optical remote sensing images,'' \emph{Int. J. Distrib.
  Sens. N.}, vol.~15, no.~5, p. 1550147719852036, 2019.

\bibitem{SGS}
D.~Zhao, J.~Wang, J.~Shi, and Z.~Jiang, ``Sparsity-guided saliency detection
  for remote sensing images,'' \emph{J. Applied Remote Sens.}, vol.~9, pp.
  1--14, 2015.

\bibitem{ROI1}
L.~Ma, B.~Du, H.~Chen, and N.~Q. Soomro, ``Region-of-interest detection via
  superpixel-to-pixel saliency analysis for remote sensing image,'' \emph{IEEE
  Geosci. Remote Sens. Lett.}, vol.~13, no.~12, pp. 1752--1756, 2016.

\bibitem{ROI2}
L.~Zhang and K.~Yang, ``Region-of-interest extraction based on frequency domain
  analysis and salient region detection for remote sensing image,''
  \emph{{IEEE} Geosci. Remote. Sens. Lett.}, vol.~11, no.~5, pp. 916--920,
  2014.

\bibitem{ROI3}
Y.~Zhang, L.~Zhang, and X.~Yu, ``Region of interest extraction based on
  multiscale visual saliency analysis for remote sensing images,'' \emph{J.
  Applied Remote Sens.}, vol.~9, pp. 1--16, 2015.

\bibitem{zhang2020rssod}
L.~Zhang and J.~Ma, ``Salient object detection based on progressively
  supervised learning for remote sensing images,'' \emph{IEEE Trans. Geosci.
  Remote Sens.}, 2021.

\bibitem{GR2}
Y.~Li and A.~Gupta, ``Beyond grids: Learning graph representations for visual
  recognition,'' in \emph{Proc. NeurIPS}, 2018, pp. 9245--9255.

\bibitem{GR7}
Y.~Liao, S.~Liu, F.~Wang, Y.~Chen, C.~Qian, and J.~Feng, ``{PPDM}: Parallel
  point detection and matching for real-time human-object interaction
  detection,'' in \emph{Proc. CVPR}, 2020, pp. 482--490.

\bibitem{GR6}
Y.~Chen, M.~Rohrbach, Z.~Yan, Y.~Shuicheng, J.~Feng, and Y.~Kalantidis,
  ``Graph-based global reasoning networks,'' in \emph{Proc. CVPR}, 2019, pp.
  433--442.

\bibitem{GR5}
M.~Henaff, J.~Bruna, and Y.~LeCun, ``Deep convolutional networks on
  graph-structured data,'' \emph{arXiv preprint arXiv:1506.05163}, 2015.

\bibitem{cbam}
S.~Woo, J.~Park, J.~Y. Lee, and I.~S. Kweon, ``{CBAM}: Convolutional block
  attention module,'' in \emph{Proc. ECCV}, 2018, pp. 1--19.

\bibitem{d3net}
D.-P. Fan, Z.~Lin, Z.~Zhang, M.~Zhu, and M.-M. Cheng, ``Rethinking {RGB-D}
  salient object detection: Models, data sets, and large-scale benchmarks,''
  vol.~32, no.~5.\hskip 1em plus 0.5em minus 0.4em\relax IEEE, 2020, pp.
  2075--2089.

\bibitem{Res2Net}
S.-H. Gao, M.-M. Cheng, K.~Zhao, X.-Y. Zhang, M.-H. Yang, and P.~Torr,
  ``{Res2Net}: A new multi-scale backbone architecture,'' \emph{IEEE Trans.
  Pattern Anal. Mach. Intell.}, 2020.

\bibitem{imagenet}
J.~Deng, W.~Dong, R.~Socher, L.-J. Li, K.~Li, and L.~Fei-Fei, ``Imagenet: A
  large-scale hierarchical image database,'' in \emph{Proc. CVPR}, 2009, pp.
  248--255.

\bibitem{Xavier}
X.~Glorot and Y.~Bengio, ``Understanding the difficulty of training deep
  feedforward neural networks,'' in \emph{Proc. AISTATS}, 2010, pp. 249--256.

\bibitem{CMC}
Z.~Liu, D.~Zhao, Z.~Shi, and Z.~Jiang, ``Unsupervised saliency model with color
  markov chain for oil tank detection,'' \emph{Remote Sens.}, vol.~11, no.~9,
  pp. 1--18, 2019.

\bibitem{COD-1}
Q.~Zhai, X.~Li, F.~Yang, C.~Chen, H.~Cheng, and D.-P. Fan, ``Mutual graph
  learning for camouflaged object detection,'' in \emph{Proc. CVPR}, 2021, pp.
  12\,997--13\,007.

\bibitem{COD-2}
H.~Mei, G.-P. Ji, Z.~Wei, X.~Yang, X.~Wei, and D.-P. Fan, ``Camouflaged object
  segmentation with distraction mining,'' in \emph{Proc. CVPR}, 2021, pp.
  8772--8781.

\end{thebibliography}
}

\end{document}